\def\year{2021}\relax
\documentclass[letterpaper]{article} 

\usepackage{amsmath,amssymb,amsfonts}
\usepackage{algorithmic}

\usepackage{textcomp}
\usepackage[FIGTOPCAP]{subfigure}
\usepackage{xcolor}
\usepackage{algorithm}
\usepackage{aaai21}  
\usepackage{times}  
\usepackage{helvet} 
\usepackage{courier}  
\usepackage[hyphens]{url}  
\usepackage{graphicx} 
\usepackage{natbib}  
\usepackage{caption} 
\newtheorem{thm}{Theorem}
\title{Uncertainty Quantification in CNN Through the Bootstrap of Convex Neural Networks}

\author{
    Hongfei Du \textsuperscript{\rm 1},
    Emre Barut \textsuperscript{\rm 2},
    Fang Jin \textsuperscript{\rm 1}
}
\affiliations{
    \textsuperscript{\rm 1} The George Washington University\\
    \textsuperscript{\rm 2} Amazon.com, Inc.\\
    hongfei@gwu.edu, ebarut@amazon.com, fangjin@gwu.edu
}

\begin{document}

\maketitle

\begin{abstract}
Despite the popularity of Convolutional Neural Networks (CNN), the problem of uncertainty quantification (UQ) of CNN has been largely overlooked. Lack of efficient UQ tools severely limits the application of CNN in certain areas, such as medicine, where prediction uncertainty is critically important. Among the few existing UQ approaches that have been proposed for deep learning, none of them has theoretical consistency that can guarantee the uncertainty quality. To address this issue, we propose a novel bootstrap based framework for the estimation of prediction uncertainty. The inference procedure we use relies on convexified neural networks to establish the theoretical consistency of bootstrap. Our approach has a significantly less computational load than its competitors, as it relies on warm-starts at each bootstrap that avoids refitting the model from scratch. We further explore a novel transfer learning method so our framework can work on arbitrary neural networks. We experimentally demonstrate our approach has a much better performance compared to other baseline CNNs and state-of-the-art methods on various image datasets.

\end{abstract}

\section{Introduction}


Artificial neural networks have been a huge success in many areas and its uncertainty quantification (UQ) is also an important task for many machine learning practitioner \cite{Krzywinski2013,Ghahramani2015}.
Lack of uncertainty
quantification can severely limit -if not completely hinder-
the deep learning applications in numerous fields.
Straightforward examples such as medicinal applications where confidence intervals are commonly used to evaluate the usefulness of
a treatment option while weighing its possible side effects; and in  deep reinforcement learning, an upper bound on the reward
of action has to be quantified to perform proper exploration.
Furthermore, as the importance of bias and fairness in machine learning
becomes mainstream, practitioners need to test hypotheses
that use model outputs, e.g., ``is gender a determining factor for
the predictions?''.  All these kind of application scenarios cannot be accomplished without a UQ framework to provide theoretical inference.



 \cite{paass1993assessing} is the first work that suggests the use of bootstrap in neural networks.
 \cite{tibshirani1996comparison} compares bootstrap with other UQ methods for neural networks, and finds bootstrap to be more ideal than other approaches. More recently, \cite{kho2015} uses bootstrap for UQ in neural networks and proposes a prediction interval optimized cost function to train the neural networks. However, all these approaches are limited due to non-convexity, that means it is not clear if the optimal solution can be obtained at every bootstrap sample, which can result in a wide confidence intervals. 

To summarize, the challenge of uncertainty quantification has the following aspects. Firstly, to make the theoretical inference, it is challenging to build a UQ framework with a proper probabilistic framework that explains the variations in the fitted models. Specifically, one needs to be able to formulate the distribution of the data generating process and the sampling distribution, which can be used to quantify the uncertainties in the fitted parameters. 
Secondly, procedures involved in training a neural network, i.e. stochastic gradient descent and its variations, are often intractable. Due to issues such as non-convexity, it is very difficult to obtain theoretical bounds on the consistency of
the predictions, or the quality of the final fitted model. 
Additionally, neural networks often contain millions of parameters and are trained on datasets with sample sizes that are on
the same order. 
This process limits the application of statistical
inference which deals with a limited number of variables and
often asymptotically infinite samples. Finally, as shown in \cite{zhang2017}, the
best performing neural networks training on CIFAR10 datasets~\cite{Krizhevsky09learningmultiple} tend to over-fitting very easily. An overfitted neural network is bound to underestimate its uncertainty on hold-out samples, and thus any approach that solely
relies on neural networks for UQ will result in over-confident estimates for uncertainty.


To overcome the above challenges, we propose a novel framework to obtain
prediction intervals by bootstrapping the predictions of convexified
convolutional neural networks (CCNN) \cite{yuchen2017} with the assistance of transfer learning method, which could further improve the framework's performance and compatibility. The contribution of this paper is threefold:
\begin{itemize}
\item Firstly, we construct a novel framework for uncertainty quantification (UQ) through Bootstrap of Convex Neural Networks, which provides the prediction intervals for uncertainty measurement. To best of our knowledge, we are the first to formulate the distribution of the data generating process and the sampling distribution, and mathematically prove that the predictions from bootstrap CCNN are asymptotically consistent, which provides solid theoretical support for our framework.

\item Secondly, we creatively integrate transfer learning with our proposed UQ framework to overcome the limitation of CCNN. Previously, CCNN could only be applied to two layers CNN. Within the combined transfer learning framework,  
we could perform UQ for arbitrary neural networks, both convex and non-convex CNN. This pioneering break-though contribution makes our bootstrap CCNN framework be able to adapt to much broader application domains. 
\item Lastly, when combining transfer learning method with CCNN, the classification accuracy and stability are better than baseline CNNs and state-of-the-art methods in various classification tasks. Extensive experiments were implemented to demonstrate this results.

\end{itemize}
The outline of this paper is as follows. In the next section, we give a brief overview of related work. After that, we explain the construction of CCNN in the section \ref{sec:ccnn}. In section \ref{sec:Boots}, we formulate our procedure and demonstrate how the method can be combined with transfer learning. In section \ref{sec:theory}, we mathematically prove the theoretical consistency properties. In section 6, we show our experimental results of our approach performance on multiple datasets. Finally, we draw our conclusion in the last section.

\section{Related Work}

One big family for UQ focus on Bayesian approaches where the uncertainty is quantified through the variation in the posterior distribution. In this context, one method models the weights from neural network with a Gaussian distribution, and obtain
the posterior using variational inference \cite{Blundell2015}. Another framework proposes the MC dropout measure, where dropout is used
to obtain the range of possible predictions \cite{Gal2016}. Their suggested procedure is computationally efficient, in the sense that the dropouts can be done after the model is trained; however, later experiments show that the estimated uncertainty does not reduce as the sample size increases \cite{osband2018}.

The other big family for UQ is to make use of ensemble methods, begin with the approach propose to quantify the uncertainty through the use of ensemble models \cite{carney1999confidence}, where multiple neural networks are trained and the uncertainty is
quantified through the difference of their predictions. More recently, one approach independently trains multiple nets and uses the variance of
the predictions as a proxy for the uncertainty \cite{Lak2017}. Another work proposes to build
an ensemble in which the samples are bootstrapped at each iteration
and a network is fit with a shrinkage penalty which forces the weights
to be similar to a target network \cite{osband2018}. They show the efficacy of their approach for deep reinforcement learning problems where the uncertainty estimates are crucial for exploration. However, ensemble methods need to train each neural network independently from scratch, so it is not computationally efficient. Moreover, ensemble methods still suffer from the non-convex problem, which ends up with inconsistent outputs from different training.  

Besides these two mainstreams, there are also other attempts for UQ. One approach uses the delta-method to build prediction intervals for feedforward networks and establish its statistical consistency \cite{hwang1997prediction}. Another method introduced a new (non-convex) loss function that depends on the range of possible predictions, which can be leveraged to compute prediction intervals for fully connected neural networks instead of a point estimate  \cite{pearce2018high}. Other work using quantile loss to produce prediction intervals in a computationally attractive manner is developed \cite{Natasa2019}. 
However, all the above approaches rely on classic neural networks, which suffer from the non-convex nature and over-fitting problem.

\section{Convex Convolutional Neural Networks} 
\label{sec:ccnn}

We provide the necessary notations and present an overview of CCNN, which are obtained through a convex relaxation of the two hidden layers CNN. A detailed version of the presentation can be found in the Appendix.

In this paper, we consider the dimension of the input image $x$ is $l_1 \times l_1 \times d$. Without loss of generality, 
we assume the image length and the width are the same and are both given by $l_{1}$. $d$ represents the number of channels. $d$ is either 1 or 3 for black\&white and color images, respectively. The label for each image $x_i$ is denoted by $y_i$ for $i=1,\dots,n$ where $n$ is the sample size. For simplicity, we further assume there are only two classes, that is $y_i \in \{0,1\}$. The extension to multiple classes can be easily obtained by following the framework in the Appendix, or by rephrasing the classification problem with a one versus all setup where a different model is built for each class \cite{Allwein2000}.

CCNN, just like any other CNN, computes classification scores from patches of the input image. For a sample $x_i$, we denote its patches as $z_p(x_i)$, where $p$ is from 1 to $P$, and $P$ is the total number of patches. We assume that the patch size is $l_2 \times l_2 \times d$, where $l_2$ corresponds to the size of the convolution kernel. We further denote the stride of the filter as $s$, and thus $P=\left(\left(l_1 - l_2 \right)/s + 1\right)^2$. Additionally, each patch $z_p(x_i)$ is vectorized and hence $z_p(x_i) \in \mathbb{R}^{d l_2^2}$. Finally, parameters in the model are $A_1,\dots,A_P$ with $A_p \in \mathbb{R}^{d l_2^2}$. Then in the framework with linear activation function, the CCNN classification score for $x_i$ is given by

\begin{equation} \label{eq:fx_linear}
f(x_i) = \sum_{p=1}^P A_p^T z_p(x_i).
\end{equation}

The terms in $A_p$ correspond to multiplication of the convolution filters and the weights between middle and the final layer. As the convolution filters should be the same for all patches, we would expect that the matrix $A=[A_1,\dots,A_P]$ is low-rank. CCNN enforces the low-rank structure by minimizing the nuclear norm of $A$, $\|A\|_{*}$, which is defined as the absolute sum of its singular values, leading to the final objective function
\begin{equation}
\min_{\|A\|_* \leq C} \sum_{i=1}^n L(f(x_i),y_i) ,
\end{equation}
 where $L$ is often the cross entropy loss function, and $C>0$ is an constant. \cite{yuchen2017} showed that with an appropriate choice of $C$, the method finds a classifier whose expected loss is bounded above with the expected loss of an optimal (non-convex) CNN. \cite{yuchen2017} also suggested using projected gradient descent to minimize the objective function due to the high efficiency. Since the projection algorithm could be executed in a stochastic fashion, so that each gradient step processes a mini-batch of examples.

The previous formulation only results in linear networks, which can be extended to networks with non-linear activation functions by the use of the kernel trick, i.e., instead of using the actual patch values, Firstly, an appropriate kernel $k(\cdot,\cdot)$ will be chosen, and the corresponding kernel matrix, $K \in \mathbb{R}^{nP \times nP}$ for each patch and sample pair will be computed. Secondly, a factorization matrix $Q$ will be computed, where $K=QQ^T$. Finally, the $z_p(x_i)$ in the original formulation is replaced with the relevant row $Q_k$ from the kernel matrix $Q$, where $Q_k$ corresponds to the $p^{\text{th}}$ patch from the $i^{\text{th}}$ sample. Thus, in the non-linear framework, the CCNN score $x_i$ is given by
\begin{equation}f(x_i) = \sum_{p=1}^P A_p^T Q(x_i,p) ,\end{equation}
where $Q(x_i,p)$ denotes the relevant row of $Q$ for sample $i$ and patch $p$, and $A_p$ is a matrix that has $nP$ many components. \cite{yuchen2017} establish that with the appropriate choice of the kernel $k(\cdot,\cdot)$, e.g., Gaussian radial kernel, this class of CCNN includes convolutional neural networks with non-linear activations for which the activation has a polynomial expansion. Unfortunately, the commonly utilized RELU is not in this collection, but the smoothed RELU is. We again refer to the Appendix for the details of each operation. Based on the convexity from CCNN, we discuss our framework in next section. 

\section{Bootstrapping with CCNN}
\label{sec:Boots}

In this section, we first discuss our proposed bootstrap CCNN framework in detail. Then, we show how the method can be extended to CNN with multiple layers using our novel transfer learning method.

\subsection{Bootstrap CCNN}

Our approach follows the classical bootstrap setup, in which the
dataset is sampled (with replacement) at each iteration. During each bootstrap, we use the parameter of the previous bootstrap $A_{b-1}$ as the initial point. With this ``warm start'' approach, we reduce the number of necessary training iterations by an order of magnitude. We note that this would not be possible if the formulation was not convex. For instance, if the warm start technique is used to train usual (non-convex) CNN, the last solution would be the local optimum closest to the previous one, and hence predictions obtained with the last bootstrap would implicitly rely on samples that are not included in its training data, canceling the statistical validity of the procedure. As the distribution of the predictions is close to the sampling distribution, the empirical distribution of the prediction probabilities provides a consistent estimate for the true probabilities. The prediction interval is then generated by the empirical bootstrap confidence interval. We also provide all the procedures in Algorithm 1.

\begin{algorithm}[tb]
    \caption{CCNN Bootstrap}
 \begin{algorithmic}
    \STATE {\bfseries Input:} Training dataset $D=\{(x_{i},y_{i})\}_{i=1}^{n}$, test dataset $T=\{(x_{i}^{'},y_{i}^{'})\}_{i=1}^{n^{'}}$ for which the prediction intervals need to be computed, confidence level $\alpha$, and the number of bootstraps $B$.
    \vspace{0.1in}
    \STATE {$A_0 \gets \arg\min L(A;D)$ \COMMENT{Train CCNN on $D$ and store fitted weights} }
       \vspace{0.1in}
    \FOR{$b$ in 1:$B$} 
    \STATE $D_b \gets \text{sample with replacement}(D,n)$ \COMMENT{Create bootstrap sample}
           \vspace{0.05in}
    \STATE $A_b \gets \arg\min L(A;D_b,A_{init}=A_{b-1})$ \COMMENT{Initialize a new CCNN model with previous weights, train on $D_{b}$}
           \vspace{0.05in}
    \STATE $pp_{b,i,k}\gets\frac{\exp(\hat{f}_{b,k}(x_{i}^{'}))}{\sum_{j=1}^{d_{2}}\exp(\hat{f}_{b,j}(x_{i}^{'}))}, for \  i=1,...,n^{'}, k=1,...,d_2$, where $d_2$ is total number of classes. \COMMENT{Compute predictions for test data}
    
    \ENDFOR
           \vspace{0.1in}
    \STATE \textbf{Calculate} $(pp_{lic},pp_{uic})$, where $pp_{lic}$ and $pp_{uic}$ are the $\frac{\alpha}{2}\%$ and $\frac{(1-\alpha)}{2}\%$ percentile of $(pp_{1,i,c},pp_{2,i,c},...,pp_{B,i,c})$ for $c\in[d_{2}]$ and $i\in[n^{'}]$  
           \vspace{0.1in}
    \STATE \textbf{Output:}  $(pp_{lic},pp_{uic})$ for $c\in[d_{2}]$ and $i\in[n^{'}]$, the $(1-\alpha)\%$ C.I.s for the predictions.
 \end{algorithmic}
\end{algorithm}

Our approach is beneficial due to two main reasons: Firstly, the convexity of the CCNN procedure guarantees the global optimum for the subsampled dataset and the statistical validity of the procedure. Secondly, for each bootstrap, we do not have to fit the model from scratch and can instead initialize the parameters to the previous
solution. This `warm-start' is possible is because of the convexity that the global optimum can be obtained regardless of the initial point. When the initial point is close to the optimum, then fewer iterations are needed, which means it saves significant computational time. 
\subsection{Transfer Learning}

The original CCNN formulation can only build neural networks with two hidden layers. \cite{yuchen2017} proposes to get around this limitation by successively adding more layers, where multiple CCNN are in the model and at each iteration convolution output from current CCNN will be passed to the next CCNN. Theoretically, this multi-stage framework is difficult to study. We instead propose to utilize transfer learning to generalize two layers CCNN for multi-layers neural networks' task.

Our approach relies on the availability of another CNN, denoted $CNN$, that has been trained for a similar task. For most image classification problems, one can use a neural network trained on ImageNet, such as VGG16 or Resnet50 \cite{He2015}. We propose to use the outputs from the last convolution layer of this network as inputs to the CCNN. The transfer learning method replaces each $x_i$ with $f_{CNN}(x_i)$ where $f_{CNN}(\cdot)$ returns the output of the last convolution layer of the deep neural network $CNN$.

We note that the transfer learning does not annul the validity of the bootstrap approach, as long as the pre-trained network used for transfer learning does not depend on any samples in our training data. Otherwise, if the pre-trained network relies on samples that are included in the training data, then the training data drawn during bootstrap would not be independent of each other. This is critical as our theoretical results are conditional on the independence of the observations.

In certain applications, such a pre-trained network may not be available. To get around this obstacle, we propose three possible approaches to create neural networks that can be used in pre-training. All of these approaches first train a CNN on our original dataset and then adjust the weights of this learned neural network so that the deep learner ``forgets'' what it has learned from the training data. We argue that the convolution filters learned with these approaches are still useful for our training data, and that the added randomness due to ``forgetting'' cancels the dependency between the outputs of the neural network and the training data, leading to a consistent bootstrap procedure. Although these techniques do not have the theoretical validity of the basic transfer learning approach, we find that they still yield comparable performances. We list these approaches below.

\begin{enumerate}
\item {\bf Train and Forget:} We train CNN on the training dataset. After a certain number of epochs, we replace the training data with an irrelevant dataset and continue training until the accuracy of the original dataset declines to a completely random guess. For instance, for classifying the MNIST dataset, one can first fit on MNIST, and then on Fashion-MNIST \cite{xiao2017/online}.
\item {\bf Train and Flip:} We train a CNN on the training dataset with the original labels for a certain number of epochs. Then, we randomly flip the labels and continue training until the deep learner learns to overfit to the random labels.
\item {\bf Train and Perturb:} After a CNN is trained, we add random perturbations to the weights of the CNN. The size of the perturbations are chosen to guarantee that the prediction of the CNN is equal to a random guess.
\end{enumerate}

\section{Theoretical Results}
\label{sec:theory}
In this section, we show that with certain modifications, the CCNN can be consistently bootstrapped. Our modified CCNN formulation minimizes the following objective function:
\begin{equation} \label{new_formulation_0}
\min_{A} \frac{1}{n}\sum_{i=1}^n L(f(x_i),y_i) + \lambda \|A\|_{*\mu},
\end{equation} where $f(x_i)$ is given by the linear form in equation \eqref{eq:fx_linear}, $\lambda>0$ is a regularization parameter and $\|A\|_{*\mu}$ is the smoothed nuclear norm:
\begin{equation}
  \|A\|_{*\mu} = \sup_{\|Z\|_2 \leq 1} \left(\text{Tr}(A^TZ) - \mu/2 \|Z\|_{Fr}^2 \right),
\end{equation} where $\|Z\|_{Fr}$, $|Z\|_2$ are given by the Frobenius norm and the spectral norm of a matrix $Z$, respectively, and $\text{Tr()}$ is the matrix trace. Furthermore, we have that $\|A\|_{*\mu} \rightarrow \|A\|_{*}$ as $\mu \rightarrow 0$. We present the results for single class classification, thus we take the loss function to be the log-loss, i.e.,
\begin{equation}L(f(x_i),y_i)=-\log(p(f(x_i))y_i - \log(1-p(f(x_i))(1-y_i),\end{equation}
where $p(z)=(1+\exp(z)^{-1})^{-1}$.
We also note that due to duality \cite{rockafellar1970convex}, the formulation in \eqref{new_formulation_0} is equivalent to
\begin{equation} \label{new_formulation}
\min_{\|A\|_{*\mu} \leq C} \sum_{i=1}^n L(f(x_i),y_i)
\end{equation}
for some constant $C>0$.

In the new formulation, the use of the smoothed nuclear norm ensures that the loss function is differentiable. Thus, the functional that maps the distribution, $F$, to the estimated function $f(x)$ is Hadamard differentiable, making bootstrap consistent \cite{van1996weak}. Next, our first theoretical result, Theorem \ref{thm:1}, states that we can consistently estimate the sampling distribution of our predictions with bootstrap. 
\begin{thm} \label{thm:1}
For a data generating process $\mathbb{P}$, let $f_{\mathbb{P},\lambda}$ be the function of form \eqref{eq:fx_linear} for which $A$ is the minimizer of 
\begin{equation}
    \min_{A} \mathbb{E}_{(X,Y) \sim \mathbb{P}} \left[ L(f(X),Y) \right]+ \lambda \|A\|_{*\mu}.
\end{equation}
Similarly, let $f_{\mathbb{P}_n,\lambda}$ be the estimated function that results from minimizing \eqref{new_formulation_0} over an empirical measure $\mathbb{P}_n$ whose observations are drawn independently from $\mathbb{P}$. For any $x$, let 
\begin{equation}
    H_n(z;x) = P \left[  f_{\mathbb{P}_n,\lambda}(x)  - f_{\mathbb{P},\lambda} (x) \leq z\right],
\end{equation}
\begin{equation}
 H_{Bn}(z;x) = P_B \left[  f_{\mathbb{P}^{B}_n,\lambda} (x)  - f_{\mathbb{P}_n,\lambda}(x)\leq z\right],   
\end{equation}
where $P_B$ is the probability over bootstraps and $\mathbb{P}^{B}_n$ refers to the empirical distribution obtained by bootstrapping from $\mathbb{P}_n$. Then, for any $x$, it holds that
\begin{equation}
  \sup_{z} \left| H_{Bn}(z;x) - H_n(z;x) \right| \rightarrow 0 \qquad \text{in probability}.  
\end{equation}
\end{thm}

The proof is provided in the Appendix. The proof relies on the work of \cite{chatterjee2005generalized} which establishes the consistency of the bootstrap for M-estimators. We note that the theorem does not guarantee the consistency of the bootstrap procedure. That is, confidence intervals from bootstrap might not be consistent if the minimizer over the data generating process, $f_{\mathbb{P},\lambda}$ is not consistent either. Our next theorem establishes that consistency can be obtained if the data are separable.

\begin{thm}
Assume that the data are separable, i.e. there is an unknown function $f^*(x)$ that can achieve perfect classification. Further, assume that the function $f^*(x)$ can be represented as a two-hidden layer CNN with linear activation functions and a finite number of convolution filters. Then, the bootstrap is consistent for the minimizer of equation~\eqref{new_formulation_0}.
\end{thm}
\begin{figure*}[ht!]
\begin{center}
\includegraphics[scale=0.65]{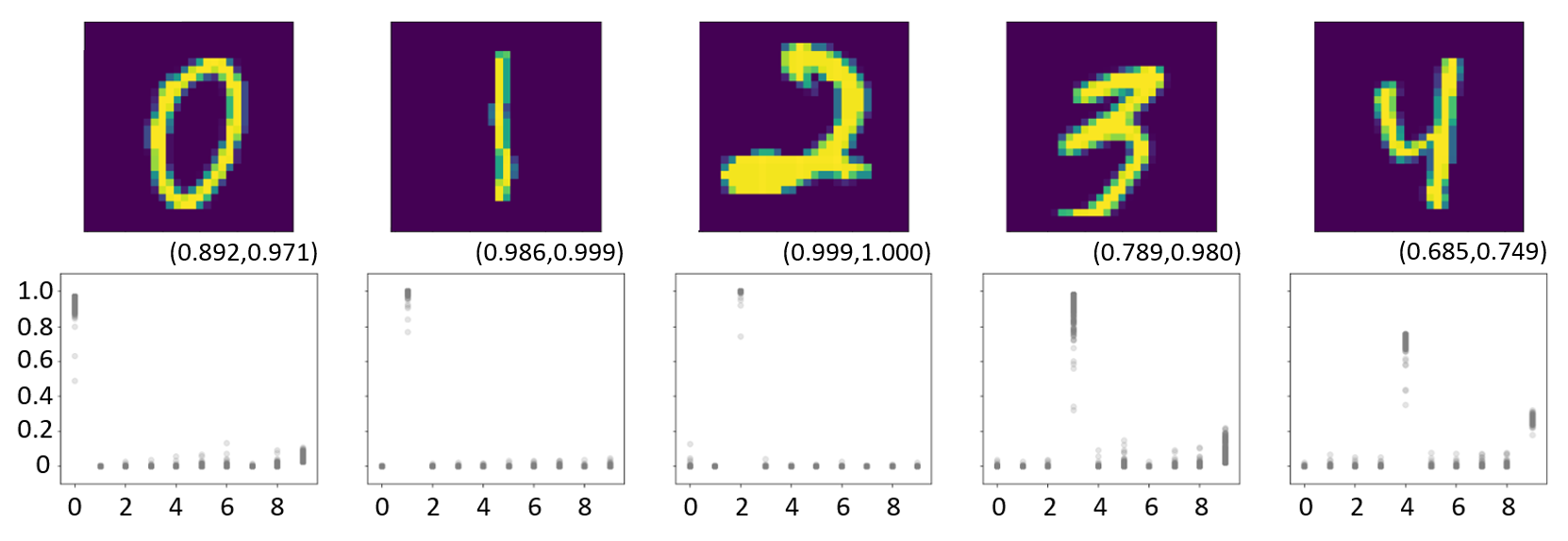}

\caption{Application of the new bootstrap approach on MNIST.
The first row displays the digit images and the distributions of the predictions
are plotted across 1000 bootstraps in the second row. The 95\%
confidence interval of the prediction probabilities for the right class
are also provided.}
\label{fig:figure1}
\end{center}
\end{figure*}

\begin{table*}[ht]
\caption{Average log-likelihood and average interval length comparison among
CCNN, ensemble method and CNN for 4 different datasets. For each dataset, first row is the average log-likelihood and second row is the average interval length. Standard errors are provided in parentheses.}

\begin{center}
{
\begin{tabular}{cccc}       
\hline
&\multicolumn{3}{c}{Average log-likelihood/ Interval length}\\ \\ \hline     & {\em CCNN} &{\em Ensemble} &{\em CNN} \\   \hline      
\\ MNIST&{\bf -3.050 (0.022)}  & -5.193 (0.022) &-6.891 (0.025) \\      &{\bf 0.0010 (0.0002)} & 0.0045 (0.0015) & 0.0021 (0.0012)\\     \hline

     \\ MNIST-blur& -8.773 {\bf (0.056)} & -10.320 (0.174)& {\bf -7.810} (0.224) \\     &  0.0074 {\bf(0.0012)} &  0.0300 (0.0033) & {\bf 0.006} (0.0019) \\     \hline

    \\ Cats and Dogs&{\bf -222.686 (0.685)}& -239.654 (0.986) & -334.797 (0.967) \\          &{\bf 0.0649 (0.0040)} & 0.129 (0.0071) & 0.0715 (0.0051) \\
  \hline  
  \\  Fashion MNIST&{\bf -355.46 (0.81)}&-363.19 (5.38)
 & -458.06 (1.90)  \\       &{\bf 0.072 (0.0047)} &0.116 (0.0074) & 0.105 (0.0074) \\     \hline

\end{tabular}}
\label{tab:table1}
  \end{center}
\end{table*}

The proof is included in the Appendix and uses the generalization bounds for CCNNs derived by \cite{yuchen2017} with adjustments specific to our case. Combining Theorem 1 and Theorem 2, we prove the consistency of our bootstrap procedure.

Few remarks are in order. Although the stated theorems are only valid for linear networks, our results still apply to CCNNs with non-linear activation functions where the ``features'' $z_p(x_i)$ are replaced by terms derived from the kernel matrix, as in Section~\ref{sec:ccnn}. However, if these features are to be obtained from the kernel matrix, there are a couple of issues that require attention: (i) bootstrap relies on the independence of observations, and the new features have to be independent of each other; (ii) the number of features, i.e. the dimension of the replacements for $z_p(x_i)$, need to be of a smaller order than $n$ as $n\rightarrow \infty$. Neither of these conditions is met with the suggested kernel framework in Section~\ref{sec:ccnn}, but they can be satisfied with minor modifications.

For the independence requirement, we propose to use a secondary dataset that has the same data generating process as the original training data and evaluate the kernel using the secondary dataset. That is, instead of calculating $K(x_i,x_j)$, one computes $K(x_i,\tilde{x}_j)$ where $\tilde{x}_j$ are from the secondary dataset. Note that the data points in the secondary dataset do not have to be labeled, and obtaining such new unlabeled data should be feasible for most image classification tasks. Another, rather dull, alternative is to set aside some portion of the training data and use it as the secondary dataset.

The requirement on the size of the features can be satisfied by fewer kernel evaluations. In practice, this is almost never an issue, as most practitioners (and algorithms) rely on approximations of the kernel matrix rather than its full form.

\section{Experiment Results}

In this section, we exhibit the results of our numerical experiment for
the bootstrap procedure discussed in section \ref{sec:Boots}. We use five datasets: MNIST \cite{Lecun1998}, noisy MNIST \cite{basu2015},
fashion MNIST \cite{xiao2017/online}, CIFAR10 \cite{Krizhevsky09learningmultiple} and the cats and dogs dataset \cite{par2012}, which contains
30,000 images of various cats and dogs. The CCNN runs on the 16 cores CPU with 64GB RAM, and other classic neural networks run on GPU.

\subsection{Demonstration on MNIST}

In our first experiment, we apply our bootstrap CCNN procedure to the
MNIST dataset \cite{Lecun1998} to obtain prediction intervals of classification outputs. As it is relatively
easy to obtain high accuracy on MNIST dataset and to better evaluate our procedure's performance of UQ when higher variation presents in the outputs, we reduce the size of the training dataset (only 1000 images in the train set and 100 images in test set) and also reduce the training iterations to 5 at each bootstrap. We set the number of bootstraps $B=1000$ and calculate the prediction intervals for the test dataset. Results are given in
Figure\ref{fig:figure1}, the distributions of the prediction probabilities for 5 randomly selected digits from the test dataset are presented and more results are included in the Appendix. From Figure\ref{fig:figure1},
we observe that for each digit, the prediction probabilities for correct digits are much higher than the wrong digits. Also,  our procedure detects higher uncertainty in classification task for digit 3 and 4 given their wider prediction intervals. To further test our procedure's performance of UQ, we experiment our procedure on various datasets and compare its performance with baseline CNNs and ensemble method in section~\ref{subsec: comparisons}.

\begin{table*}[ht]
\centering
\caption{Average log-likelihood and average interval length comparison among three transfer learning approaches, CCNN and the ensemble method with 20 nets. For each dataset, first row is the average log-likelihood and second row is the average interval length. Standard errors are provided in parentheses.}

{
\begin{tabular}{cccccc}     
\hline
&\multicolumn{5}{c}{Average log-likelihood/Interval length} \\ \\ 
\hline  &{\em Ensemble} &{\em CCNN}   & {\em Forget} & {\em Flip}&{\em Perturb}  \\         
\hline \\ Fashion MNIST&-363.19 (5.38) &-355.46 (0.81) &{\bf-261.187} (0.298) & -461.285 (0.359) &-445.957 {\bf(0.285)} \\     &0.116 (0.0074)&0.072 (0.0047) &{\bf 0.0703} (0.0031)&	0.0934 	(0.0031) &	0.0925 {\bf(0.0025)} \\     \hline

  \\  CIFAR10 & -666.082 (6.205)& -576.417 (0.533)&{\bf-525.736}	(0.532) &-577.382	(0.436) & -555.675	{\bf(0.324)}  \\    &0.576 (0.01)&0.168 (0.003) &0.170	(0.004)  &0.166	(0.003)& {\bf0.138	(0.002)}\\     \hline 

\end{tabular}}
 \label{tab:table2}
\end{table*}

\subsection{Comparisons Versus Alternatives}
\label{subsec: comparisons}

In this subsection, we compare our procedure with two alternative techniques: (i) the ensemble method with 20 nets \cite{Lak2017}; and (ii) bootstrap with non-convex CNN. In each experiment, we evaluate each method's accuracy and uncertainty on test datasets using two criteria:
\begin{enumerate}
\item Interval length: Given as the average length of the 95\%
confidence interval. Shorter interval length is more preferred, which indicates lower uncertainty. 
\item Average log-likelihood: Given as the average log-likelihood of the
observations over the estimates of prediction probabilities. More specifically,
the score is given by
\begin{equation}
  L=\frac{1}{B}\sum_{b=1}^{B}\sum_{i=1}^{N}H\left(p_{i}^{b},y_{i}\right),  
\end{equation}

where $p_{i}^{b}$ is the estimated probabilities for sample $i$
in bootstrap $b$, $y_{i}$ is the one-hot encoding for the true class
and $H(\cdot,\cdot)$ is the cross-entropy. In this metric, larger scores are more preferable as they suggest less entropy between the modeled distribution and the outcome, which also indicates the smaller difference between $p_{i}^{b}$ and $y_{i}$ over all test samples and bootstraps. Therefore, higher average log-likelihood implies model's higher overall prediction accuracy.

\end{enumerate}

Our experiment uses the following datasets:
\begin{enumerate}
\item MNIST \cite{Lecun1998} with 10 classes of handwritten digits. The images' size is 28x28 and in gray scale. We use 60,000 images for training and 1,000 images for testing.
\item Noisy MNIST  \cite{basu2015} with motion blur added to original MNIST dataset. The images' size and sizes of training and testing datasets are same as above.
\item Fashion MNIST Dataset containing 10 classes of clothes \cite{xiao2017/online}. The images' size and sizes of training and testing datasets are same as above.
\item Cats and Dogs \cite{par2012}. The images' size is 224x224x3 and in RGB. We use 10,000 images for training and 1,000 images for testing.
\end{enumerate}

For the first three datasets, the ensemble method and the bootstrap CNN use the classic CNN, Le-Net, with 3 convolution and 2 fully connected layers, where the numbers of convolution filters are (32,64,128) with a kernel size of (2,2). 

CCNN uses the standard two-hidden layers setup for MNIST and Noisy MNIST. For training CCNN on the Fashion MNIST dataset, we utilize transfer learning. Firstly, we train a CNN on original MNIST and then use it as the pre-trained model. Next, we feed the Fashion MNIST dataset to the pre-trained model and the outputs from its last convolution layer are used as the input for CCNN.

For the Cats and Dogs dataset, we use transfer learning for all methods and utilize VGG16 \cite{sim2015}. For ensemble and bootstrap CNN, we use transfer learning to retrain the last three layers of VGG16. For CCNN, we the feed Cats and Dogs dataset to the pre-trained VGG16 and outputs from its last convolution layer are used as input for CCNN. Both CCNN and CNN use 100 bootstraps. The ensemble method uses 20 networks (VGG16). We provide the results of our experiments in Table~\ref{tab:table1} and {we could observe that our bootstrap CCNN approach yields higher log-likelihood and shorter intervals on average for 3 datasets as shown in bold numbers, which demonstrates the higher prediction accuracy and lower uncertainty. Also, the corresponding standard errors in the parentheses are smaller than the other two methods, which shows our approach provides more stable predictions and more consistent uncertainty measurement. For the MNIST-blur dataset, our method achieves similar levels of accuracy (average log-likelihood) and uncertainty (average interval length) as the baseline CNN method, while our method has smaller standard errors in both cases, which shows that under comparable performance, our method provides more stable prediction and more consistent uncertainty measurement. The above observations support our theoretical claim that the non-convex neural networks (CNN) tend to have higher uncertainty in predictions due to the difficulty of convergence to global optimal. Moreover, our bootstrap UQ method could detect this higher uncertainty, which suggests that our approach can be used reliably to quantify uncertainty in complex machine vision tasks.}

\subsection{Comparison of Transfer Learning Approaches}

We generalize the two layers CCNN by three novel transfer learning approaches, which are listed in Section~\ref{sec:Boots}. In this subsection, We compare their performance on two datasets: Fashion MNIST \cite{xiao2017/online} and CIFAR10 \cite{Krizhevsky09learningmultiple}.

The pre-trained networks for transfer learning use the same architecture as in the previous subsection, with 3 convolution layers and 2 fully connected layers. The pre-trained networks are built in the following manner:

\begin{enumerate}
    \item {\bf Train and Forget:}  We train the CNN on Fashion MNIST data (cats and dogs from CIFAR10) for 30 epochs. Then, the same network is trained on Original MNIST data (deer and horse from CIFAR10) for another 30 epochs. 
    
    \item{\bf Train and Flip:} We train the CNN on Fashion MNIST data (cats and dogs from CIFAR10) for 30 epochs. Then, we train the CNN on the same datasets with randomly flipped labels for another 30 epochs.
    
    \item{\bf Train and Perturb:}  We train the CNN on Fashion MNIST data (cats and dogs from CIFAR10) for 30 epochs. Then, we add random Gaussian perturbations (with $\sigma=0.5$ and $\sigma=0.1$ for Fashion MNIST and CIFAR10, respectively) to all of the weights. After the perturbation, accuracy of the model is very close to 10\% for Fashion MNIST experiment (50\% for CIFAR10 experiment).
    
\end{enumerate}

After the pre-trained CNN is built, the outputs of its last convolution layer are used as the input for the CCNN model. Then, the prediction intervals are estimated with 100 bootstraps. We explore different tuning parameters for CCNN and the best performances in terms of the average log-likelihood and the average interval length are given in Table~\ref{tab:table2}. We provide the experiment results for all of the settings we considered in the Appendix. We also find that the performance difference between the best and the worst hyper-parameters are not significant.

From Table~\ref{tab:table2}, we find that the proposed transfer learning methods provide an efficient tool for generalizing the CCNN framework for multiple layers networks' task, as evidenced by the shorter interval lengths and the higher average log-likelihoods. We also observe that `Train and Perturb' method has the smallest standard errors for average log-likelihood and interval length in both experiments, which may serve as a conservative choice. ``Train and Forget'' method has the best overall performance in terms of higher accuracy (average log-likelihood) and lower uncertainty (average interval length). In the meantime, it also achieves similar levels of standard errors as ``Train and Perturb'' method. To summary, we conclude the overall best performing transfer learning approach is ``Train and Forget'', which consistently outperforms the ensemble method and original CNN method as shown in Table~\ref{tab:table2}.

\section{Conclusion}

Due to the non-convex nature of CNN, it is hard to guarantee the outputs converge to global optimal results, making the uncertainty quantification of CNN a challenging task. To solve this problem, we propose a novel framework which combines the bootstrap method and CCNN in this paper. We prove that our approach can consistently estimate the sampling distribution of the predictions with the bootstrap method, and thus provides theoretical support for our framework. Moreover, we explore an innovative transfer learning method, `Train and Forget', to improve convexified neural networks' prediction accuracy and reduces its uncertainty, which also enables our framework works for arbitrary neural networks. Our experimental results show our proposed bootstrap CCNN framework combined with the `Train and Forget' transfer learning method achieves better accuracy and stability compared to the baseline CNNs and state-of-the-art methods. 


\bibliography{bibfile}

@article{Lecun1998,
	Author = {Y. LeCun and L. Bottou and Y. Bengio and P. Haffner},
	Journal = {Proceedings of the IEEE},
	Title = {Gradient-based learning applied to document recognition},
	Volume={86},
	Pages={2278-2324},
	Year = 1998}

@article{Blundell2015,
	Author = {Charles Blundell and Julien Cornebise and Koray Kavukcuoglu and Daan Wierstra},
	Journal = {International Conference on Machine Learning},
	Title = {Weight Uncertainty in Neural Networks},
	Volume={37},
	Pages={1613-1622},
	Year = 2015}

@article{Gal2016,
	Author = {Yarin Gal and Zoubin Ghahramani},
	Journal = {International Conference on Machine Learning},
	Title = {Dropout as a Bayesian Approximation: Representing Model Uncertainty in Deep Learning},
	Volume={48},
	Year = 2016}

@article{osband2018,
	Author = {Ian Osband and John Aslanides and Albin Cassirer},
	Journal = {Conference on Neural Information Processing Systems},
	Title = {Randomized Prior Functions for Deep Reinforcement Learning},
	Year = 2018}

@article{Lak2017,
	Author = {Balaji Lakshminarayanan and Alexander Pritzel and Charles Blundell},
	Journal = {Conference on Neural Information Processing Systems},
	Title = {Simple and Scalable Predictive Uncertainty Estimation using Deep Ensembles},
	Year = 2017}

@article{zhang2017,
	Author = {Chiyuan Zhang and Samy Bengio and Moritz Hardt and Benjamin Recht and Oriol Vinyals},
	Journal = {International Conference on Learning Representations},
	Title = {Understanding Deep Learning Requires Rethinking Generalization},
	Year = 2017}

@article{kho2015,
	Author = {A. Khosravi and S. Nahavandi and D. Srinivasan and R. Khosravi},
	Journal = {IEEE Transactions on Neural Networks and Learning Systems},
	Title = {Constructing Optimal Prediction Intervals by Using Neural Networks and Bootstrap Method},
	Year = 2015}

@article{yuchen2017,
	Author = {Yuchen Zhang and Percy Liang and Martin J. Wainwright},
	Journal = {International Conference on Machine Learning},
	Title = {Convexified Convolutional Neural Networks},
	Year = 2017}

@article{basu2015,
	Author = {Saikat Basu and Manohar Karki and Sangram Ganguly and Robert DiBiano and Supratik Mukhopadhyay and Ramakrishna Nemani},
	Journal = {Neural Process Lett},
	Title = {Learning Sparse Feature Representations using Probabilistic Quadtrees and Deep Belief Nets},
	Volume={45},
	Pages={855-867},
	Year = 2017}

@article{par2012,
	Author = {O. M. Parkhi and A. Vedaldi and A. Zisserman and C. V. Jawahar},
	Journal = {IEEE Conference on Computer Vision and Pattern Recognition},
	Title = {Cats and Dogs},
	Year = 2012}

@article{sim2015,
	Author = {K. Simonyan and A. Zisserman},
	Journal = {International Conference on Learning Representations},
	Title = {Very deep convolutional networks for large-scale image recognition},
	Year = 2015}

@misc{xiao2017/online,
    title={Fashion-MNIST: a Novel Image Dataset for Benchmarking Machine Learning Algorithms},
    author={Han Xiao and Kashif Rasul and Roland Vollgraf},
    year={2017},
    eprint={1708.07747},
    archivePrefix={arXiv},
    primaryClass={cs.LG}
}

@article{Krizhevsky09learningmultiple,
    author = {Alex Krizhevsky},
    title = {Learning multiple layers of features from tiny images},
    note = {Technical report},
    year = {2009}
}

@article{He2015,
	author = {Kaiming He and Xiangyu Zhang and Shaoqing Ren and Jian Sun},
	title = {Deep Residual Learning for Image Recognition},
	journal = {IEEE Conference on Computer Vision and Pattern Recognition},
	year = {2016}
}

@article{Natasa2019,
	author = {Natasa Tagasovska and David Lopez-Paz},
	title = {Single-Model Uncertainties for Deep Learning},
	journal = {Conference on Neural Information Processing Systems},
	year = {2019}
}

@incollection{van1996weak,
  title={Weak convergence},
  author={Van Der Vaart Aad W and Wellner Jon A},
  booktitle={Weak convergence and empirical processes},
  pages={16--28},
  year={1996},
  publisher={Springer}
}

@book{rockafellar1970convex,
  title={Convex analysis},
  author={Ralph Tyrell Rockafellar},
  number={28},
  year={1970},
  publisher={Princeton university press}
}

@article{chatterjee2005generalized,
  title={Generalized bootstrap for estimating equations},
  author={Chatterjee Snigdhansu and Bose Arup},
  journal={The Annals of Statistics},
  volume={33},
  number={1},
  pages={414--436},
  year={2005},
  publisher={Institute of Mathematical Statistics}
}

@inproceedings{paass1993assessing,
  title={Assessing and improving neural network predictions by the bootstrap algorithm},
  author={Paass and Gerhard},
  booktitle={Conference on Neural Information Processing Systems},
  pages={196--203},
  year={1993}
}

@article{tibshirani1996comparison,
  title={A comparison of some error estimates for neural network models},
  author={Tibshirani Robert},
  journal={Neural Computation},
  volume={8},
  number={1},
  pages={152--163},
  year={1996},
  publisher={MIT Press}
}

@article{hwang1997prediction,
  title={Prediction intervals for artificial neural networks},
  author={Hwang JT Gene and Ding A Adam},
  journal={Journal of the American Statistical Association},
  volume={92},
  number={438},
  pages={748--757},
  year={1997},
  publisher={Taylor \& Francis}
}

@inproceedings{carney1999confidence,
  title={Confidence and prediction intervals for neural network ensembles},
  author={Carney John G and Cunningham P{\'a}draig and Bhagwan Umesh},
  booktitle={International Joint Conference on Neural Networks. Proceedings (Cat. No. 99CH36339)},
  volume={2},
  pages={1215--1218},
  year={1999},
  organization={Institute of Electrical and Electronics Engineers}
}

@inproceedings{pearce2018high,
  title={High-quality prediction intervals for deep learning: A distribution-free, ensembled approach},
  author={Pearce T and Zaki M and Brintrup A and Neely A},
  booktitle={International Conference on Machine Learning},
  volume={9},
  pages={6473--6482},
  year={2018}
}

@article{Allwein2000,
title={Reducing multi-class to binary: A unifying approach for margin classifiers},
author={E. L. Allwein and R. E. Schapire and Y. Singer},
journal={Journal of Machine Learning Research},
volume={1},
pages={113--141},
year={2000}

}

@article{Krzywinski2013,
title={Points of significance: Importance of being uncertain},
author={M Krzywinski and N Altman},
journal={Nature methods},
volume={10},
pages={9},
year={2013}

}

@article{Ghahramani2015,
title={ Probabilistic machine learning and artificial intelligence},
author={Z Ghahramani},
journal={Nature},
volume={521},
pages={7553},
year={2015}

}

\end{document}